\documentclass{article}
\usepackage{ijcai21}

\usepackage{times}
\usepackage{soul}
\usepackage{url}
\usepackage[hidelinks]{hyperref}
\usepackage[utf8]{inputenc}
\usepackage[small]{caption}
\usepackage{graphicx}
\usepackage{amsmath}
\usepackage{amsthm}
\usepackage{booktabs}
\usepackage{algorithm}
\usepackage{algorithmic}
\title{Investigating the Relationship Between World Development Indicators and the Occurence of Disease Outbreaks in the 21\textsuperscript{th} Century: A Case Study}

\author{
    Content Areas:
    \affiliations
    Paraphrase Identification \\
    Semantic Equivalence \\
    Transformer Architecture
}

\author{
Aboli Marathe$^1$\thanks{Equal Contributors}
\and
Harsh Sakhrani$^1$\footnotemark[1]\And
Saloni Parekh$^1$\footnotemark[1]
\affiliations
$^1$Pune Institute of Computer Technology, Maharashtra, India\\
\emails
\{aboli.rajan.marathe, harshsakhrani26, saloniparekh1609\}@gmail.com,
}

\begin{document}

\maketitle

\begin{abstract}
  The timely identification of socio-economic sectors vulnerable to a disease outbreak presents an important challenge to the civic authorities and healthcare workers interested in outbreak mitigation measures. This problem was traditionally solved by studying the aberrances in small-scale healthcare data. In this paper, we leverage data driven models to determine the relationship between the trends of World Development Indicators and occurrence of disease outbreaks using worldwide historical data from 2000-2019, and treat it as a classic supervised classification problem. CART based feature selection was employed in an unorthodox fashion to determine the covariates getting affected by the disease outbreak, thus giving the most vulnerable sectors. The result involves a comprehensive analysis of different classification algorithms and is indicative of the relationship between the disease outbreak occurrence and the magnitudes of various development indicators. 
\end{abstract}

\section{Introduction}
\par Coronavirus (COVID-19) has become an unprecedented health crisis and has spread to over 150 countries, severely impacting the world economy and causing social disruption. In a recent study from 2020, it was observed that the COVID-19 outbreak had a  significant impact on the Italian economy, eventually tipping it into recession. The impact of this recession fell on the financially weak population, elderly and the working population. ~\cite{a1}  Learning from our experiences, we wish to move forward and create robust emergency preparedness measures. Providing the authorities with the most vulnerable sectors which will succumb to disease outbreaks first will be an invaluable resource for planning and policy-making. But finding these vulnerable sectors presents a challenge as it requires big data analysis of imperfect data over multiple years of history for a particular region. Furthermore, the vulnerable sectors cannot be directly quantified, thus their vulnerability needs to be estimated through indirect measures. We studied such cases of previous disease outbreaks, and propose a method of accurately identifying these vulnerable sectors, including critical sectors like economy, healthcare and safety. We came across the World Development Indicators (WDI), established by the World Bank that are a set of indicators, collected over time for every country through their individual governments. The indicators cover most sectors of development, including trade and safety markers and we thought of using these indicators to estimate the vulnerability of different sectors. Some world development indicators get more affected than others and their identification is a challenging problem for disease outbreak preparedness and planning. 
\par Over the years, researchers have analysed the disease outbreaks to determine the risk factors ~\cite{a2} and aid the disease outbreak surveillance ~\cite{a3}. But the relationship between trends in socio-economic indicators and the occurrence of previous disease outbreaks still remains a mystery. While researchers tend to analyse socio-economic systems in the context of disease outbreaks, we tried understanding the relationship between socio-economic systems and disease outbreaks. Whether metadata could be used to  analyze the anomalies and their cause or impacts on a country-per-country basis was our primary research question. We tried to answer this question using  a different methodology. In this paper, we propose an approach that uses data-driven models for disease outbreak identification rather than disease outbreak forecasting and the treatment of this identification as a classification problem is a novel approach that we would like to introduce to this field. We combine the World Development Indicators data ~\cite{a4,a5} provided by the World Bank and the disease outbreaks data by the World Health Organization ~\cite{a6} to create a dataset. As there was a small degree of uncertainty in the dataset due to missing values, we also make use of statistical data imputation and predictive modelling for data treatment. Lastly, we apply benchmarked classification techniques for disease outbreak identification and CART based feature importance to find the crucial indicators. After finding these indicators, we compare the results with former studies and surveys to validate the performance of this methodology. The verified indicators can be passed on to the authorities for emergency preparedness and planning assistance.

\section{Background Work}
Due to the recent rise of disease outbreaks, the research community is laying special emphasis on studying epidemiology. Researchers have found correlation of time-series data trends with the presence of disease outbreaks ~\cite{a7,a8} and have also found causal relationships between socio-cultural systems ~\cite{a9}. ~\cite{a10} worked extensively to model these outbreaks and predict them. The case-study of Salmonella agona in their paper highlighted both the potential and the shortcomings of automated detection procedures, emphasising both their time optimization and less perceptible results. ~\cite{a11} tried another approach, using  hierarchical time series analysis model to detect outbreaks and found the proposed model to be a reliable tool for Rubella notifications and Salmonella infections.  ~\cite{a12} considered continuous-time stochastic compartmental models that can be applied in veterinary epidemiology to model the within-herd dynamics of infectious diseases. ~\cite{a13} introduced a statistical method for detection of specific types of aberrations in public health surveillance. 

~\cite{a14} summarised the work of multiple authors in an attempt to identify the secondary impacts of these disease outbreaks in certain countries. They analyse the economic, political, social and secondary impacts of the outbreaks, unlike the traditional healthcare impacts and found some common features among the countries struck by outbreaks. We were inspired by these results and wondered if one methodology, applied on single or multiple datasets could reproduce these findings in sufficiently timely fashion to allow interventions to take place.  While most studies are trying to identify the impact of disease outbreaks using statistical modelling, few try to analyse this problem in a reverse manner, i.e. socio-economic indicators that could be associated with the occurrence of a disease outbreak. The directionality of this indicator-outbreak network has been considered a problem too vast for any single study, something that we agree with but look forward to solving. In 2011 however, ~\cite{a15} discussed a wide variety of techniques, their possible limitations and advantages from regression to ARIMA models to Markov models for the identification of unusual patterns in data which may result from infectious disease outbreaks. This study provided a base for our methodology.

\section{Data Description and Creation}

World Development Indicators (WDI) is the primary World Bank collection of 143 development indicators for more than 200 economies and 40 country groups. The part of the database that we considered spans from the year 2000 - 2019. ~\cite{a5} The disease outbreak data from WHO was extracted separately for individual countries. ~\cite{a6} The years that had a disease outbreak occurrence/absence were labelled as 1/0 respectively. 

The basic preprocessing involved encoding categorical features, scaling, normalization and resampling. Robust Scaler was utilized for scaling, since it scales the data according to the quantile range and is insensitive to outliers. A number of other scaling techniques like Min-Max scaler, Standard scaler and our in-house Logarithmic Deviation scaler were also tried, but gave substandard results. 

The severely skewed class distribution observed in our dataset posed a challenge for the classification algorithms. Both undersampling and oversampling have known disadvantages. Undersampling can throw away potentially useful data, and oversampling can increase the likelihood of overfitting. Hence, a combination of both Undersampling and Oversampling was used. SMOTE is an oversampling technique that synthesizes new plausible examples of the minority class by interpolating between several minority class examples that lie together. Tomek Links refers to an undersampling technique that identifies cross-class nearest neighbors and removes the majority class occurrence. ~\cite{a16}

\begin{figure*}
    \centering
    \includegraphics[width=\textwidth]{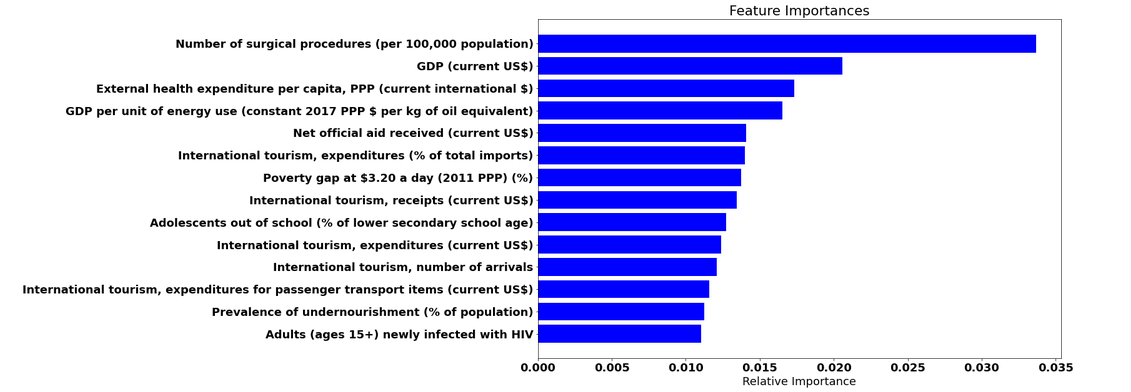}
    \caption{Feature importance top covariates}\label{fig:tikz1}
\end{figure*}

\section{Methodology}

\subsection{Data Imputation}
We employed a number of statistical and inferential data imputation techniques ranging from simple statistic substitution to complex deep learning based imputation techniques. The techniques that gave us noteworthy results are explained below.

\subsubsection{KNN imputation}
The K-Nearest Neighbors algorithm is used to map a point with its k closest neighbors in a multi-dimensional space. The intuition behind using KNN for data imputation is that a missing input variable value can be approximated by the value of the points that are closest to it, and this ‘closeness’ can be determined on the basis of other non-missing variables. In our dataset, the missing World Development Indicator values are imputed using this 'closeness', which is usually seen in groups of countries having similar indicator values, or the countries that have had similar development curves in different time frames. After experimenting with a number of parameters, the best results were obtained when a combination of 5 neighbors and euclidean distance \eqref{eq:1} was used.
\begin{equation}
    d\left( p,q\right)   = \sqrt {\sum _{i=1}^{n}  \left( q_{i}-p_{i}\right)^2}
    \label{eq:1}
\end{equation}

\subsubsection{Stochastic Multiple Regression Imputation}
The intuition behind the MSREG algorithm is to leverage the correlation between the input variables by regressing the missing variable on all the other input variables. We employ the Linear Regression Model to estimate the missing values. For example, in our dataset there is a strong positive correlation between the “Number of Community Health Workers” and the “Current health expenditure” columns. The MSREG algorithm is capable of utilizing such correlations in order to impute the missing variable values.

To counter the decrease in the inherent variability of the imputed variable, normally distributed noise with a mean of zero and variance equal to the standard error of regression estimates was introduced. The MSREG method assigns values to each missing element $x$ according to \eqref{eq:2}, where $k$ is the number of manifest variables used in a model, $N$ is the number of missing values in $x$, and $Srandn()$ is a function that returns a different element of a standardized normally distributed random column vector each time it is invoked. ~\cite{a17}

\begin{equation}
\dot{x}_{ir}  =  \sum_{j=1}^{k}  \hat{\beta}_{x_{i}x_{j}} + ( \sqrt{(1-\hat{\beta}_{x_{i}x_{j}} \hat { \Sigma }_{x_{i}x_{j}}} ) \text{Srandn()}
\label{eq:2}
\end{equation}

{\centering where $j = 1 ... k, j$ $\neq$ $i, r = 1 ... N$}

\subsubsection{Random (Additive Noise) Imputation}
This method allows imputation of the missing data by picking random observed values of a particular variable. This method was applied on all features with missing data, by selecting random variables with the probability of an imputation being $1/n$ where $n$ is the number of present values.

\subsection{Feature Importance}
Conventionally Feature Selection has always been used to identify the relevant set of features for which there is a significant increase in the performance of the algorithm. But we attempt to utilize it in an unorthodox fashion. The promising classification scores [\ref{sample-table}] do show that there is a strong correlation between the World Development Indicators and Disease Outbreaks. But the crucial question would be to discover the set of unapparent Indicators which get affected by Disease Outbreaks and understand them better, for which we use Feature Selection.

RandomForestClassifier’s implicit feature selection was used to determine the subset of relevant features. For randomized trees' ensembles, the variable importance $X_m$ for predicting $Y$ is calculated by adding up the weighted impurity decreases $p(t)\Delta i(st, t)$ for all nodes $t$ where $X_m$ is used, averaged over all $N_T$ trees in the forest \eqref{eq:3}.

\begin{equation}
Imp(X_m) =  \frac{ 1 }{ N_{T}} \sum_{T}^{} \sum_{t \epsilon T;v(s_{t}) = X_{m}}^{} p(t) \Delta i(s_{t},t) 
\label{eq:3}
\end{equation}

where $p(t)$ is the proportion $N_t/N$ of samples reaching $t$ and $v(s_t)$ is the variable used in split $s_t$. When using the Gini index as an impurity function, this measure is known as the Gini importance or Mean Decrease Gini. ~\cite{a18}

\begin{table*}[h]
  \caption{Results of classification algorithms on the three imputed datasets}
  \label{sample-table}
  \centering
  \begin{tabular}{lcccccc}
    \toprule
    Algorithms &\multicolumn{2}{c}{KNN} &\multicolumn{2}{c}{Random} &\multicolumn{2}{c}{MSREG}              \\
    \cmidrule(r){2-3} \cmidrule(r){4-5} \cmidrule(r){6-7}
    &F1     & Accuracy &F1     & Accuracy &F1     & Accuracy\\
    \midrule
    LGBM Classifier &0.934 &0.934 &0.918 &0.918 &0.935 &0.935\\
    RandomForest &0.930 &0.930 &0.940 &0.940 &0.942 &0.942\\
    BaggingClassifier &0.906 &0.906 &0.906 &0.906 &0.913 &0.914\\
    DecisionTreeClassifier &0.832 &0.832 &0.808 &0.808 &0.816 &0.817\\
    ANN {1 Relu , 1 Sigmoid layer} &0.810 &0.743 &0.780 &0.808 &0.780 &0.799\\
    AdaboostClassifiier &0.799 &0.799 &0.835 &0.835 &0.812 &0.812\\
    VotingClassifier &0.789 &0.790 &0.780 &0.780 &0.786 &0.786\\
    ANN {3 Relu , 1 Sigmoid layer} &0.780 &0.824 &0.780 &0.800 &0.770 &0.808\\
    Autoencoder + Logistic Reg. &0.760 &0.832 &0.780 &0.811 &0.760 &0.829\\
    Autoencoder + Logistic Reg. &0.760 &0.834 &0.790 &0.817 &0.760 &0.831\\
    GaussianNB &0.644 &0.652 &0.654 &0.662 &0.575 &0.609\\
    SVC &0.580 &0.570 &0.570 &0.523 &0.490 &0.540\\
    SGDClassifier &0.546 &0.608 &0.586 &0.617 &0.786 &0.851\\
    \bottomrule
  \end{tabular}
\end{table*}

\section{Results}
The identification of important features from the imputed dataset was accomplished through the use of benchmarked classification techniques to predict the target variable y, which in our case is the disease outbreak occurrence in a particular year. We applied these techniques to our dataset with a 0.2 train-test-split, and compared 3 different methods of imputation: KNN, Random Imputation and MSREG. After analysing the F1-score and the accuracy, hyperparameter optimization was performed to boost our results. This was followed by the usage of the CART based feature selection technique to determine the important features.

\paragraph{Classification:} A wide range of \emph{state-of-the-art} classification techniques were employed including Bayesian, Tree-based, Ensemble and Deep Learning Algorithms [\ref{sample-table}].

\paragraph{Feature selection:} We applied feature selection for filtering out the input variables strongly correlated with disease outbreak occurrence. By plotting the relative importance of these covariates, we can increase the interpretability of this pipeline, and thus deliver the vulnerable indicators as our final result [\ref{fig:tikz1}].

\section{Synopsis}
The results were very promising on the imperfect data classification as we achieved 94.2\% top accuracy and a F1-Score of 0.94 on the dataset using the Random Forest algorithm, MSREG imputation and SMOTE Sampling after hyperparameter tuning. The ensemble techniques performed better than both the regression and the deep learning models. We were able to extract the most important features that the algorithm predicted [\ref{fig:tikz1}].

\begin{figure*}
 \centering
    \includegraphics[width=\textwidth]{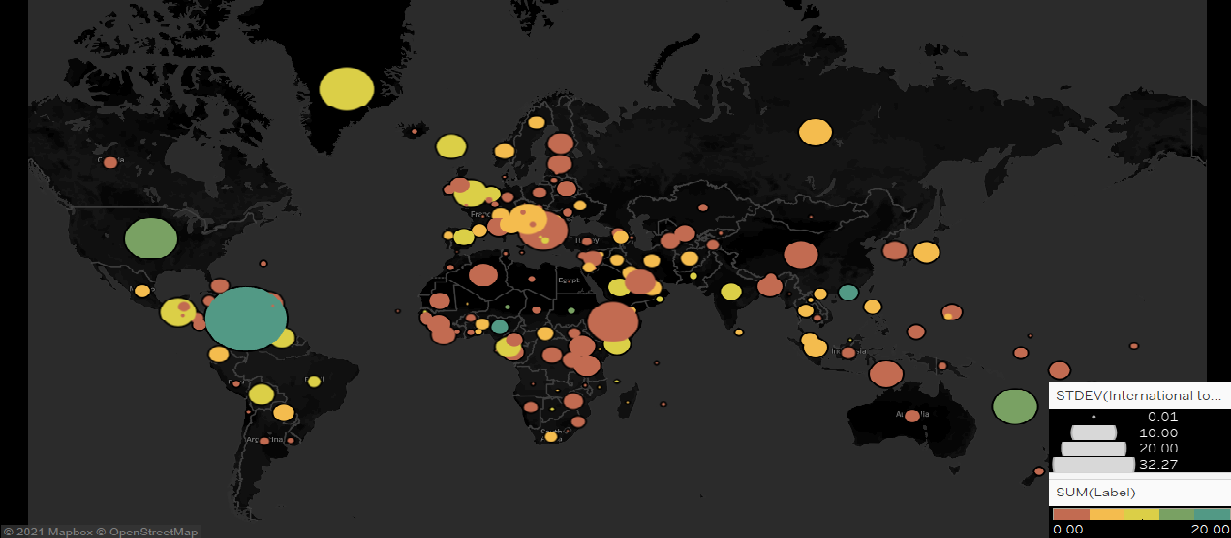}
    \caption{Standard deviation of normalised number of International tourism, expenditures (current US\$) with the frequency of disease outbreaks per country} 
    \label{figx}
\end{figure*}

To interpret our results better, we visualised the frequency of disease outbreaks per country with one of the more important predicted features- Number of International tourism, expenditures (current US\$), and observed that the two variables were indeed correlated [\ref{figx}]. It is interesting to note how our observations match the results put forward by ~\cite{a14}, where they found the above features to be strongly affected by disease outbreaks in low and middle income countries through a different methodology and dataset.

\section{Conclusion and Future Work}
The dire impact of disease outbreaks are unequivocally faced by the most vulnerable populations, the healthcare workers and the financially disadvantaged, and our insights could help the authorities increase the accessibility of social services. Our proposed method leverages data-driven models and feature selection for the quick identification of the affected indicators, giving the vulnerable sectors. The results on the imputed datasets, while indicative of potential relationships, cannot tell the whole story on their own. Many critical variables (e.g. competing political priorities, cultural narratives etc.) cannot be completely captured in a large scale analysis, and can be found by comparing public opinion and conflict-related casualties. In the future work, these insights can contribute to forming prior knowledge for a knowledge-driven model, providing concrete parameters to help assess and validate the theoretical framework. Our team has also conducted a study parallel to this work that builds on the dataset and analyzes causal relationships between the features \cite{miaox}.This research will be useful in the emergency preparedness planning for the developing world.

\nocite{a19}
\nocite{a20}
\nocite{a21}
\nocite{a22}
\nocite{a23}
\nocite{a24}
\nocite{a25}
\nocite{a26}
\nocite{a27}
\nocite{a28}
\nocite{a29}
\nocite{a30}
\nocite{a31}

\bibliographystyle{named}
\bibliography{ijcai21}

\begin{thebibliography}{}

\bibitem[\protect\citeauthoryear{Allard}{1998}]{a3}
Rbc Allard.
\newblock Use of time-series analysis in infectious disease surveillance.
\newblock {\em Bulletin of the World Health Organization}, 76(4):327, 1998.

\bibitem[\protect\citeauthoryear{Anno \bgroup \em et al.\egroup }{2019}]{a2}
Sumiko Anno, Takeshi Hara, Hiroki Kai, Ming-An Lee, Yi~Chang, Kei Oyoshi,
  Yousei Mizukami, and Takeo Tadono.
\newblock Spatiotemporal dengue fever hotspots associated with climatic factors
  in taiwan including outbreak predictions based on machine-learning.
\newblock {\em Geospatial Health}, 14(2), 2019.

\bibitem[\protect\citeauthoryear{Bank}{}]{a5}
World Bank.
\newblock World development indicators.
\newblock Accessed Feb. 14, 2021.

\bibitem[\protect\citeauthoryear{Bank}{2010}]{a4}
World Bank.
\newblock {\em World development indicators 2010}.
\newblock The World Bank, 2010.

\bibitem[\protect\citeauthoryear{Batista \bgroup \em et al.\egroup
  }{2003}]{a16}
Gustavo~EAPA Batista, Ana~LC Bazzan, Maria~Carolina Monard, et~al.
\newblock Balancing training data for automated annotation of keywords: a case
  study.
\newblock In {\em WOB}, pages 10--18, 2003.

\bibitem[\protect\citeauthoryear{Dauda}{2019}]{a23}
Rasaki~Stephen Dauda.
\newblock Hiv/aids and economic growth: Evidence from west africa.
\newblock {\em The International journal of health planning and management},
  34(1):324--337, 2019.

\bibitem[\protect\citeauthoryear{Davis \bgroup \em et al.\egroup }{2019}]{a9}
Paul~K Davis, Angela O'Mahony, and Jonathan Pfautz.
\newblock {\em Social-Behavioral Modeling for Complex Systems}.
\newblock John Wiley \& Sons, 2019.

\bibitem[\protect\citeauthoryear{Farrington and Beale}{1998}]{a10}
CP~Farrington and AD~Beale.
\newblock The detection of outbreaks of infectious disease.
\newblock In {\em Geomed’97}, pages 97--117. Springer, 1998.

\bibitem[\protect\citeauthoryear{Farrington \bgroup \em et al.\egroup
  }{1996}]{a31}
CP~Farrington, Nick~J Andrews, AD~Beale, and MA~Catchpole.
\newblock A statistical algorithm for the early detection of outbreaks of
  infectious disease.
\newblock {\em Journal of the Royal Statistical Society: Series A (Statistics
  in Society)}, 159(3):547--563, 1996.

\bibitem[\protect\citeauthoryear{Heisterkamp \bgroup \em et al.\egroup
  }{2006}]{a11}
Simon~H Heisterkamp, Arnold~LM Dekkers, and Janneke~CM Heijne.
\newblock Automated detection of infectious disease outbreaks: hierarchical
  time series models.
\newblock {\em Statistics in Medicine}, 25(24):4179--4196, 2006.

\bibitem[\protect\citeauthoryear{Kalibala \bgroup \em et al.\egroup
  }{2012}]{a20}
Samuel Kalibala, Katie~D Schenk, Deborah~C Weiss, and Lynne Elson.
\newblock Examining dimensions of vulnerability among children in uganda.
\newblock {\em Psychology, health \& medicine}, 17(3):295--310, 2012.

\bibitem[\protect\citeauthoryear{Kock}{2014}]{a17}
Ned Kock.
\newblock Single missing data imputation in pls-sem.
\newblock {\em Lar. Tex. Scr. Syst}, 2014.

\bibitem[\protect\citeauthoryear{Kuang \bgroup \em et al.\egroup }{2012}]{a27}
Jie Kuang, Wei~Zhong Yang, Ding~Lun Zhou, Zhong~Jie Li, and Ya~Jia Lan.
\newblock Epidemic features affecting the performance of outbreak detection
  algorithms.
\newblock {\em BMC Public Health}, 12(1):1--9, 2012.

\bibitem[\protect\citeauthoryear{Li \bgroup \em et al.\egroup }{2012}]{a8}
Zhongjie Li, Shengjie Lai, David~L Buckeridge, Honglong Zhang, Yajia Lan, and
  Weizhong Yang.
\newblock Adjusting outbreak detection algorithms for surveillance during
  epidemic and non-epidemic periods.
\newblock {\em Journal of the American Medical Informatics Association},
  19(e1):e51--e53, 2012.

\bibitem[\protect\citeauthoryear{Li}{2018}]{a28}
Marie Li.
\newblock Missing value estimation algorithms on cluster and representativeness
  preservation of gene expression microarray data.
\newblock {\em arXiv preprint arXiv:1809.05969}, 2018.

\bibitem[\protect\citeauthoryear{Lieberman}{2007}]{a19}
Evan~S Lieberman.
\newblock Ethnic politics, risk, and policy-making: A cross-national
  statistical analysis of government responses to hiv/aids.
\newblock {\em Comparative political studies}, 40(12):1407--1432, 2007.

\bibitem[\protect\citeauthoryear{Louppe \bgroup \em et al.\egroup }{2013}]{a18}
Gilles Louppe, Louis Wehenkel, Antonio Sutera, and Pierre Geurts.
\newblock Understanding variable importances in forests of randomized trees.
\newblock {\em Advances in neural information processing systems 26}, 2013.

\bibitem[\protect\citeauthoryear{Lu \bgroup \em et al.\egroup }{2009}]{a30}
Hsin-Min Lu, Daniel Zeng, and Hsinchun Chen.
\newblock Prospective infectious disease outbreak detection using markov
  switching models.
\newblock {\em IEEE Transactions on Knowledge and Data Engineering},
  22(4):565--577, 2009.

\bibitem[\protect\citeauthoryear{Marathe \bgroup \em et al.\egroup
  }{2021}]{miaox}
Aboli Marathe, Saloni Parekh, and Harsh Sakhrani.
\newblock Modelling major disease outbreaks in the 21st century: A causal
  approach.
\newblock 2021.

\bibitem[\protect\citeauthoryear{Novakovic and Veljovic}{2011}]{a26}
J~Novakovic and A~Veljovic.
\newblock C-support vector classification: Selection of kernel and parameters
  in medical diagnosis.
\newblock In {\em 2011 IEEE 9th international symposium on intelligent systems
  and informatics}, pages 465--470. IEEE, 2011.

\bibitem[\protect\citeauthoryear{Organization}{}]{a6}
World~Health Organization.
\newblock Disease outbreaks by countries, territories and areas.
\newblock Accessed Feb. 14, 2021.

\bibitem[\protect\citeauthoryear{Richardson \bgroup \em et al.\egroup
  }{2016}]{a7}
Eugene~T Richardson, Mohamed~Bailor Barrie, J~Daniel Kelly, Yusupha Dibba,
  Songor Koedoyoma, and Paul~E Farmer.
\newblock Biosocial approaches to the 2013-2016 ebola pandemic.
\newblock {\em Health and human rights}, 18(1):115, 2016.

\bibitem[\protect\citeauthoryear{Rohwerder}{2020}]{a14}
Brigitte Rohwerder.
\newblock Secondary impacts of major disease outbreaks in low-and middle income
  countries.
\newblock 2020.

\bibitem[\protect\citeauthoryear{Sahri \bgroup \em et al.\egroup }{2014}]{a29}
Zahriah~Binti Sahri, Universiti~Tekonologi Malaysia, et~al.
\newblock Support vector machine-based fault diagnosis of power transformer
  using k nearest-neighbor imputed dga dataset.
\newblock {\em Journal of Computer and Communications}, 2(09):22, 2014.

\bibitem[\protect\citeauthoryear{Saiya and Scime}{2019}]{a21}
Nilay Saiya and Anthony Scime.
\newblock Comparing classification trees to discern patterns of terrorism.
\newblock {\em Social Science Quarterly}, 100(4):1420--1444, 2019.

\bibitem[\protect\citeauthoryear{Sanfelici}{2020}]{a1}
Mara Sanfelici.
\newblock The italian response to the covid-19 crisis: Lessons learned and
  future direction in social development.
\newblock {\em The International Journal of Community and Social Development},
  2(2):191--210, 2020.

\bibitem[\protect\citeauthoryear{Sawers \bgroup \em et al.\egroup }{2008}]{a25}
Larry Sawers, Eileen Stillwaggon, and Tom Hertz.
\newblock Cofactor infections and hiv epidemics in developing countries:
  implications for treatment.
\newblock {\em AIDS care}, 20(4):488--494, 2008.

\bibitem[\protect\citeauthoryear{Streftaris and Gibson}{2004}]{a12}
George Streftaris and Gavin~J Gibson.
\newblock Bayesian inference for stochastic epidemics in closed populations.
\newblock {\em Statistical Modelling}, 4(1):63--75, 2004.

\bibitem[\protect\citeauthoryear{Stroup \bgroup \em et al.\egroup }{1993}]{a13}
Donna~F Stroup, Melinda Wharton, Karen Kafadar, and Andrew~G Dean.
\newblock Evaluation of a method for detecting aberrations in public health
  surveillance data.
\newblock {\em American journal of epidemiology}, 137(3):373--380, 1993.

\bibitem[\protect\citeauthoryear{Ukpolo}{2004}]{a22}
Victor Ukpolo.
\newblock Aids epidemic and economic growth: testing for causality.
\newblock {\em Journal of Asian and African Studies}, 39(3):169--178, 2004.

\bibitem[\protect\citeauthoryear{Unkel \bgroup \em et al.\egroup }{2012}]{a15}
Steffen Unkel, C~Paddy Farrington, Paul~H Garthwaite, Chris Robertson, and Nick
  Andrews.
\newblock Statistical methods for the prospective detection of infectious
  disease outbreaks: a review.
\newblock {\em Journal of the Royal Statistical Society: Series A (Statistics
  in Society)}, 175(1):49--82, 2012.

\bibitem[\protect\citeauthoryear{Vogli \bgroup \em et al.\egroup }{2014}]{a24}
Roberto~De Vogli, Anne Kouvonen, Marko Elovainio, and Michael Marmot.
\newblock Economic globalization, inequality and body mass index: a
  cross-national analysis of 127 countries.
\newblock {\em Critical Public Health}, 24(1):7--21, 2014.

\end{thebibliography}

\end{document}